\documentclass[conf]{new-aiaa}
\usepackage[utf8]{inputenc}
\usepackage{titlesec}
\usepackage{graphicx}
\usepackage{indentfirst}
\usepackage{caption}
\usepackage{amsmath}
\usepackage[version=4]{mhchem}
\usepackage{siunitx}
\usepackage{subcaption}
\usepackage{longtable,tabularx}
\usepackage[toc,page]{appendix}
\titlespacing*{\section}
{0pt}{0pt}{3.3ex}
\titlespacing*{\subsection}
{0pt}{0pt}{2.3ex plus .2ex}
\title{\LARGE CompText: Visualizing, Comparing \& Understanding Text Corpus}
\author { Suvi Varshney\footnote{MS in CS at UC Davis, suvarshney@ucdavis.edu }, Divjeet Singh Jas\footnote{MS in CS at UC Davis, djas@ucdavis.edu}}
\begin{document}
\maketitle

\section{Introduction}
A common practice in Natural Language Processing (NLP) is to visualize the text corpus without reading through the entire literature, still grasping the central idea and key points described. For a long time, researchers focused on extracting topics from the text and visualizing them based on their relative significance in the corpus. However, recently, researchers started coming up with more complex systems that not only expose the topics of the corpus but also words closely related to the topic to give users a holistic view. These detailed visualizations spawned research on comparing text corpora based on their visualization. Topics are often compared to idealize the difference between corpora. However, to capture greater semantics from different corpora, researchers have started to compare texts based on the sentiment of the topics related to the text. This gives users an idea about the weightage of a particular topic in a corpus. Comparing the words carrying the most weightage, we can get an idea about the important topics for corpus.

There are multiple existing texts comparing methods present that compare topics rather than sentiments but we feel that focusing on sentiment carrying words would better compare the two corpora. Since only sentiments can explain the real feeling of the text and not just the topic, topics without sentiments are just nouns. We aim to differentiate the corpus with a focus on sentiment, as opposed to comparing all the words appearing in the two corpora. The rationale behind this is, that the two corpora do not many have identical words for side-by-side comparison, so comparing the sentiment words gives us an idea of how the corpora are appealing to the emotions of the reader. We can argue that the entropy or the unexpectedness and divergence of topics should also be of importance and help us to identify key pivot points and the importance of certain topics in the corpus alongside relative sentiment.

For the proof of concept, we theorized that the Presidential speech dataset \cite{dataset}  would be a good fit since politicians appeal to the public emotions, hence we expect to see dramatic contrast between the two opposing ideologies.

Our \textbf{contributions} are:
\begin{enumerate}
\item We present a pipeline, CompText, for systematic analysis of two text corpora, with a focus on sentiments.
\item We implement our proof of concept on the Presidential speech dataset
\item We provide multiple case studies and use cases of our approach
\end{enumerate}

A lot of work has been presented on comparing text corpora, our novelty is in the fact that we analyze the document based on sentiment. We also prove the superiority of our results compared to the present methods in the case studies section. We believe that our work can be used in various fields like understanding how the media of different countries present the same news, can be used by brands for their competitive research \& marketing, analyzing call center data to improve efficiency.

\section{Related Work}
The sequential link between words is extremely important in the research of text sentiment analysis. Mikolov et al. \cite{Mikolov} introduced the \textit{RNN} language model, which is widely acknowledged as being capable of processing text sequence data. However, in reality, \textit{RNN} was unable to properly learn the knowledge. When the gap between relative text information and the current location to be predicted widens, various issues emerge. The backpropagation through time optimization technique \textit{(BPTT)} has too many unfold layers, which causes historical information loss and gradient attenuation during training. To address this issue, several researchers proposed the \textit{LSTM} method \cite{LSTM}, which improves experimental outcomes in particular application settings. When the text sequence information is fairly lengthy, the \textit{RNN} model with \textit{LSTM} is often more successful in overcoming the sequence information attenuation problem. So, for text sentiment analysis, we'd use an \textit{RNN} with \textit{LSTM}. In a document, \textit{N-grams} are repeated sequences of words, symbols, or tokens. They are the adjacent sequences of things in a document, in technical terminology. They are used in NLP activities when dealing with text data.

To emphasize discriminating words, a variety of visualizations have been employed, including basic ranked lists of words, word clouds, word bubbles, and word-based scatter plots. There are a variety of methods for generating word scores for ranking that has been comprehensively discussed in prior work. For an overview of model-based term scoring methods, the reader might consult Monroe et al. \cite{3}. Bitvai and Cohn \cite{bitvai} provide a method for identifying sparse words and phrase scores using a trained \textit{ANN} (with bag-of-words features) and its training data. Coppersmith and Kelly \cite{Coppersmith} propose a word-cloud-based visualization for discriminating words, although it is intended for tiny subsets of a much larger corpus. A third, middle cloud is included for phrases that appear to be distinctive. Bostock et al. \cite{bostock} use interactive word-bubble visualization to compare Republican and Democratic words used throughout the 2012 US presidential nominating conventions.

\section{Design and Methodology}
\subsection{Theoretical Concepts}
The technical description of the project is given according to the different views that we present for the comparison of the different corpus. First we define metrics that we use to compare the text. 
\begin{enumerate}
\item Entropy: Entropy accounts for both a keyword’s relative frequency and its unexpectedness. If we let $P$ denote the entire normalized distribution of words in a text with vocabulary $\mathcal{T}$, then the (Shannon) entropy \cite{shannon} is given by the equation \ref{shanon}.
\begin{equation} \label{shanon}
	\mathcal{H}(P) = {\sum\limits_{\tau \in \mathcal{T}} p_{\tau}  \log_2\frac{1}{p_{\tau}}}
\end{equation}
\item Divergence: Divergence is an asymmetric measure of how texts differ from each other. If we want to compare a reference text and a comparison text, where let $P^{(1)}$ be the relative word frequency distribution of the reference text and $P^{(2)}$ be the distribution of the comparison, then the Kullback-Leibler divergence (KLD), or relative entropy, is given by the equation \ref{div}.
\begin{equation} \label{div}
	\mathcal{D^{(KL)}}(P^{(2)}||P^{(1)}) = {\sum\limits_{\tau \in \mathcal{T}} p_{\tau}^{(2)}  \log_2\frac{1}{p_{\tau}^{(1)}} - p_{\tau}^{(2)}  \log_2\frac{1}{p_{\tau}^{(2)}}}
\end{equation}
\end{enumerate}
Next, the different views for text comparison are described as follows: 
\begin{enumerate}
    \item Word Cloud (Force Directed Graph): With this type of graph we aim to find the top most frequent sentiment carrying words in a corpus and link all the common words between different corporas. In this graph the node represents the different corpus and the edges represent the link between them.  
    The user will be able to see the top most important labeled sentiment carrying words in the corpus and also the common words used by different corporas.
    \item Word Shift Graphs:
    \begin{enumerate}
        \item Frequency comparison word shift graph:
We take the most important words for every corpora and find 30 common words to plot their relative frequencies on the word shift graphs. If $p^{(1)}_i$ is the relative frequency of word $i$ in the first text, and $p^{(2)}_i$ is its relative frequency in the second text, then the proportion shift $\delta p_{i}$ calculates their difference using $\delta p_{i}=p^{(2)}_i-p^{(2)}_i$ . If the difference is positive ($\delta p_{i}$>0), then the word is relatively more common in the second text. If it is negative ($\delta p_{i}$<0), then it is relatively more common in the first text. There is also a subplot of the cumulative contribution on the bottom left, which traces how $\sum\limits_{i}|\delta p_{i}|$ changes as we add more words according to their rank. The horizontal line shows the cutoff of the word contributions that are plotted versus those that are not. This helps show how much of the overall difference between the texts is explained by the top contributing words. 
\\
\item Entropy comparison word shift graph:
We use the Shannon entropy, $\mathcal{H}(p)$, of words to identify more surprising words and how they vary between two texts. The less often a word appears in a text, the more surprising it is. The Shannon entropy can be understood as the average surprise of a text. We can compare two texts by taking the difference between their entropies, $\delta\mathcal{H} = \mathcal{H}(p^{2})-\mathcal{H}(p^{1})$. We plot the top 30 common words with the most Shannon entropy. We also note that the absolute Shanon entropy does not mean anything, it's only the relative Shannon entropy that defines the rarity of a word. The top bar pf the graph shows the cumulative sum of $\delta\mathcal{H}$ and tells us which of the two texts are more unpredictable.
\\
\item Divergence comparison word shift graph:
The Kullback-Leibler Divergence (KLD) is a useful asymmetric measure of how two corpora/texts differ. We have a reference text and a comparison text. The KL divergence of the comparison text with respect to the reference text helps us identify the relative entropy of a word from the comparison text with respect to the reference text. We again take the common words and calculate their divergence score, plotting the word-shift graph.\\
    \end{enumerate}
\end{enumerate}
To summarize, we compare the texts focusing on the sentiment. Through the word cloud graph, the user can view the top most labeled sentiment carrying words, for the corpus between different text corpus. This graph also shows the common link in the texts by linking the most important words. Further for in-depth evaluation, we draw the word shift graphs to analyze common, most important sentiment words based on the frequency, entropy, and divergence of two text corpora.
\begin{figure}[!hbt]
     \centering
     \includegraphics[width=1\linewidth]{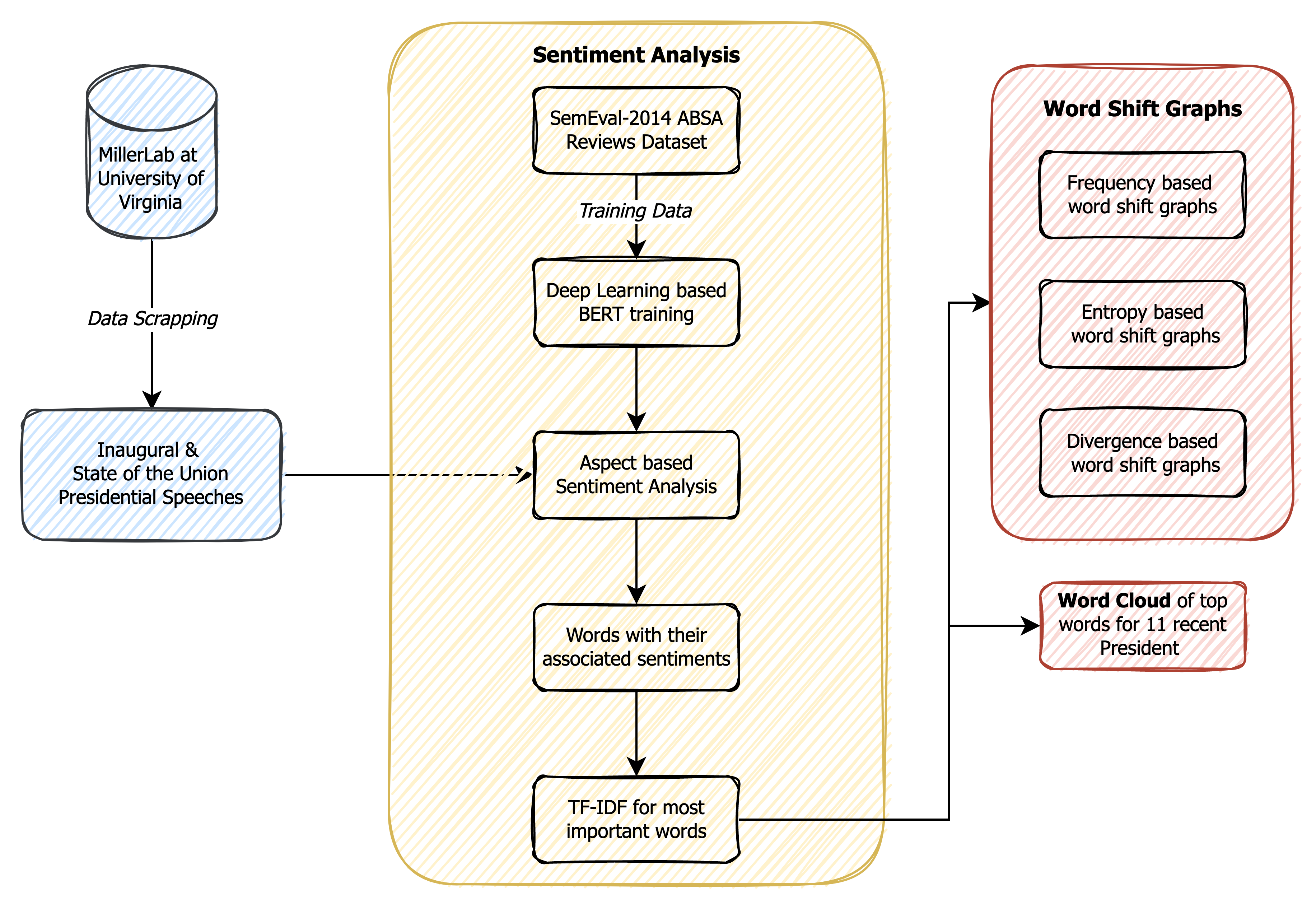}
     \caption{Model Architecture}\label{Fig:Model}
\end{figure}
\subsection{System Architecture}
There has been ample research on Aspect Based sentiment analysis. In the recent time, BERT based sentiment analysis have been topping the accuracy charts. We follow Sun et. al. \cite{sun-etal-2019-utilizing} and further train their Deep Learning model on \cite{dataset-senti} to suit our needs. Similarly, there are ample methods to draw the word cloud graph.  The field of word shift graphs is quite nascent. There have been a lot of explorations on what word shift graphs can represent. We follow the Gallagher et. al \cite{Shifterator} to build the word shift graphs. We organize these two analysis methods into a \textit{\textbf{novel}} architecture. We hypothesise that comparison on the sentiment carrying words should be able to provide the user with a detailed and meaningful difference between two corpus. The system architecture design for our project is given in the Figure \ref{Fig:Model}.\\

\subsection{Implementation}

\subsubsection{Dataset}
We describe 2 datasets in our project, one being the SemEval 2014 \cite{dataset-senti} Restaurant reviews dataset for training the sentiment analysis method. This dataset has the reviews with the sentiment carrying words and their sentiment. For the second dataset \cite{dataset}, we collected the speeches in the Inauguration and State of the Union address and appended all of them, one after the other for each president. We have the Inauguration and the State of the Union speeches for 43 presidents and the speeches are on average 20000 words long. This dataset is the primary testing dataset, although much more regressive testing has also been done. This data set contains the name of the Presidents, their speeches, and the source of their speeches. 

\subsubsection{ML Method Used}
We use transfer learning to train our sentiment analysis method. The entire code is written in python. We use the \textit{Pytorch} deep learning framework to build and train the model as well as handle the dataset. The BERT model itself is made using the \textit{NN} submodule of PyTorch. We also use the \textit{JSON} module to parse the dataset and Pandas to transform it. To review sentences from the dataset, we also employ some of the NLP modules like \textit{tokenizer}, and \textit{NLTK}.

\subsubsection{Visual Interface}
\begin{figure}[!hbt]
     \centering
     \includegraphics[width=1\linewidth]{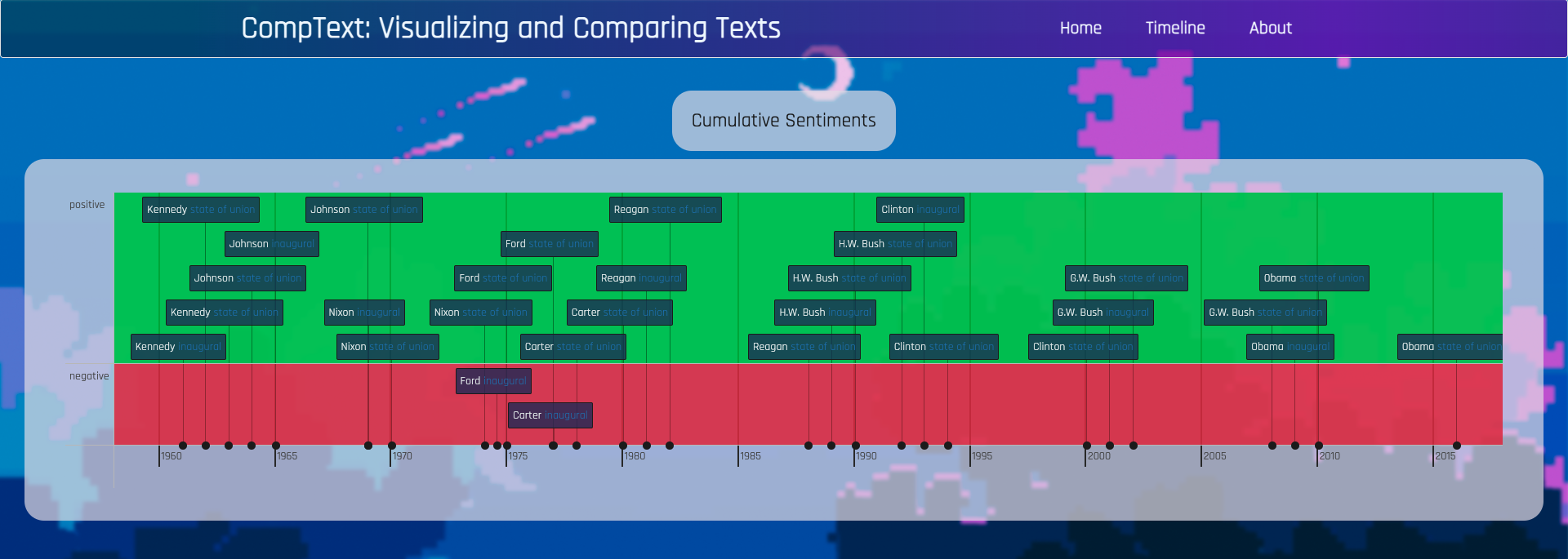}
     \caption{Cumulative Sentiments}\label{Fig:Frontend}
    \end{figure}

\begin{figure}[!hbt]
     \centering
     \includegraphics[width=1\linewidth]{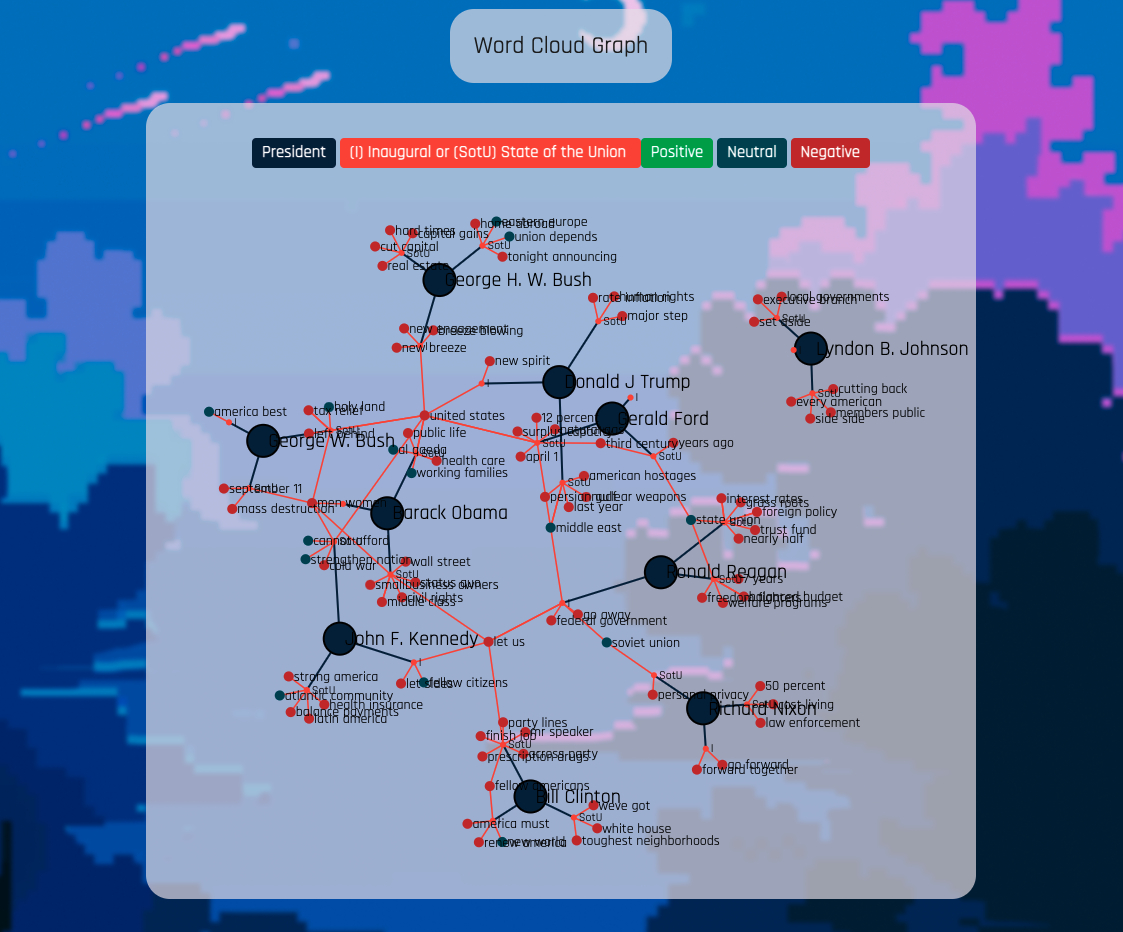}
     \caption{Word Cloud Graph}\label{Fig:Wordcloud}
    \end{figure}

We used flask, a Python-based microframework to deploy our machine learning model and make our interactive web application, because of the features it provides. For generating the forced directed graphs we have used the D3.js library. This graph is shown in Figure \ref{Fig:Wordcloud}. It shows the top most common labeled sentiment carrying words of different presidents and which presidents had spoken the same sentiment-carrying words. We have separated the speeches into overall negative and positive sentiment and visualized it using a timeline as shown in Figure  \ref{Fig:Frontend} that shows the cumulative sentiments of all the presidential speeches

\section{Evaluation and Results}
In this section, we provide detailed analysis/case studies on different text corpora through our pipeline. Since our focus is mainly on the sentiment carrying words, it is in our interest to present the analysis which are rich in sentiment. We present the improved analysis of presidential speeches and academic journal reviews.

\subsection{Case Studies}
\subsubsection{Barack H. Obama v/s Donald J. Trump}
Analysing the speeches of the presidents, the first task is to analyse all the words and record their sentiment. Through these sentiment carrying words, the user can look at the different analysis and get insights about the content of the corpora.
\begin{figure}
\centering
\begin{subfigure}{.33\textwidth}
  \centering
  \includegraphics[width=1\linewidth]{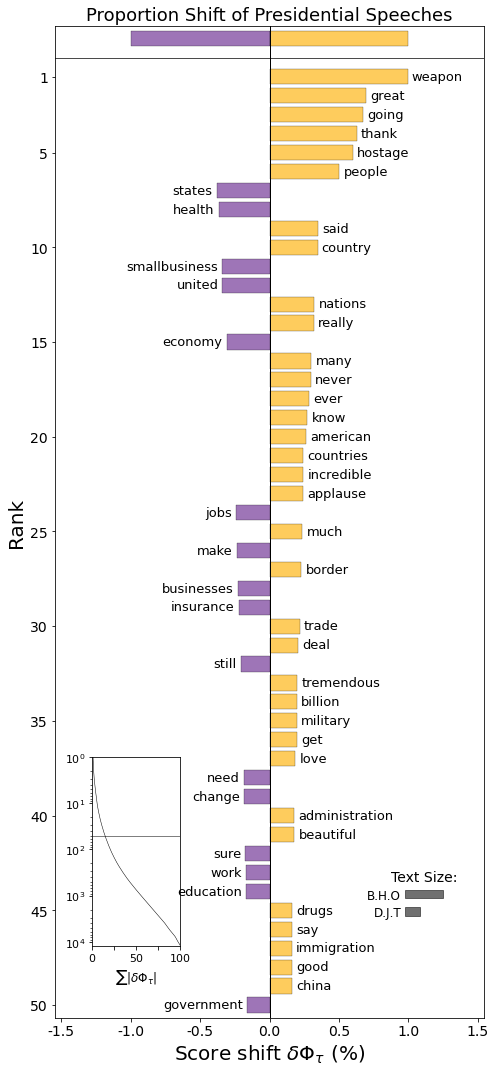}
  \caption{Frequency Comparision}
  \label{fig:freq}
\end{subfigure}%
\begin{subfigure}{.33\textwidth}
  \centering
  \includegraphics[width=1\linewidth]{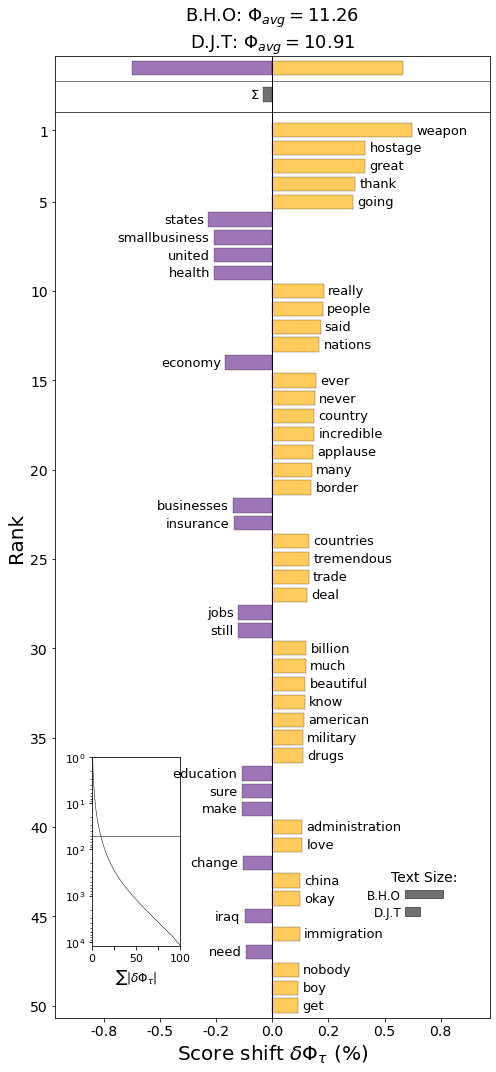}
  \caption{Entropy Comparision}
  \label{fig:entropy}
\end{subfigure}
\begin{subfigure}{.33\textwidth}
  \centering
  \includegraphics[width=1\linewidth]{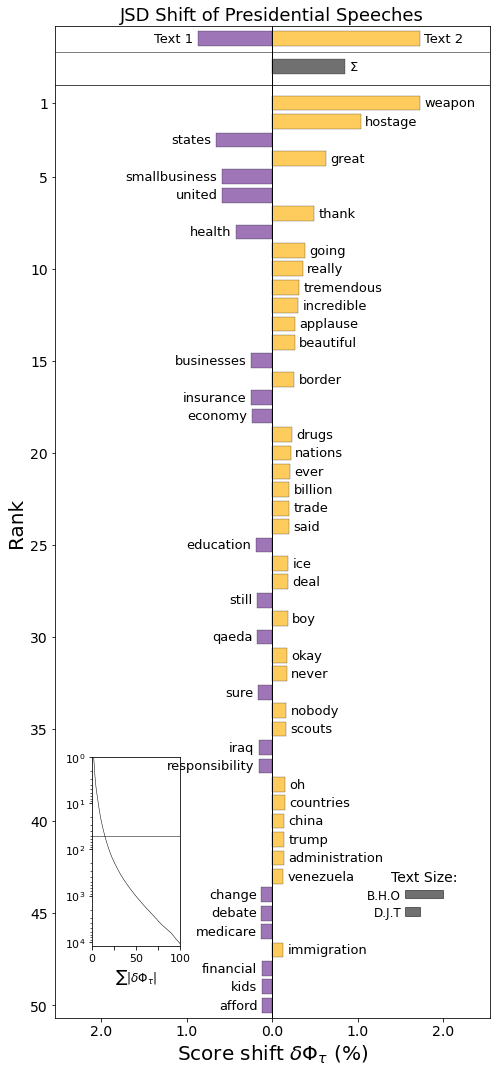}
  \caption{Divergence Comparison}
  \label{fig:kld}
\end{subfigure}
\caption{Word Shift Graphs for Presidential Speeches}
\label{fig:graphs}
\end{figure}
\begin{enumerate}
    \item Word Cloud (Force Directed Graph): Since the World Cloud graph can handle more than one corpus, it helps us to explore the connections of the different document (President speeches in this case). We can look at the words with the top aggregate sentiment in the speeches and also see the common words. In Figure \ref{Fig:Wordcloud}, we can see the topmost frequent words used by the presidents in Inaugural or State of the Union speeches. From the graph we see that President Trump and President Obama used the word \textit{'united states'} in a negative context. We also saw that Lyndon B. Johnson had no common words with any of the presidents. Through this graph, we can decide what two corpora (speeches) to compare with the word shift graphs. We also have an auxiliary view \ref{Fig:Frontend}, which helps us to get an overview of the sentiment of all the entire corpora.

    \item Word Shift Graphs:
    \begin{enumerate}
        \item Frequency comparison word shift graph:
       As mentioned above, we take the most important words for every president and find 25 common words to plot their relative frequencies on the word shift graphs. In Figure \ref{fig:freq}, the comparison of the speeches on Barack H. Obama (B.H.O in purple) and Donald J. Trump (D.J.T in yellow) is given. The first thing we notice is the corpus size, which is twice as big for President Obama as it is for President Trump, which is expected due to the tenure of both presidents. We also notice that in spite of the smaller corpora, President trump repeats the words more than President Obama, and hence his speech is less comparatively diverse.  We can see that President Trump used the word \textit{‘Billions’} more than President Obama, while President Obama used the word \textit{‘Education’}  more than President Trump. The cumulative contribution plot on the bottom left traces how $\sum\limits_{i}|\delta p_{i}|$ changes as we add more words according to their rank. The horizontal line shows the cutoff of the word contributions that are plotted versus those that are not. This helps show how much of the overall difference between the texts is explained by the top contributing words. In this plot, about a quarter of the overall difference is explained by the top 25 words. The bars on the bottom right show the relative corpora size of the presidents.
\\
\item Entropy comparison word shift graph:
In Figure \ref{fig:entropy}, we can see which words are relatively surprising in which president's speech. The words \textit{‘Great’}, \textit{‘Thank’}  are much more surprising, while the words, \textit{‘China’}, \textit{‘Border’} are much less surprising in President Trump’s Speech compared to President Obama’s speech. The top bar shows the cumulative sum of $\delta\mathcal{H}$ and tells us that President Trump's speech is a lot more unpredictable than President Obama’s speech.
\\
\item Divergence comparison word shift graph: In Figure \ref{fig:kld}, we can see that the word \textit{‘Great’} in President Trump’s speeches is surprising with respect to the reference speeches of President Obama. Similarly, the word \textit{‘Health’} in President Obama’s speeches is surprising compared to the reference, President Trump’s speeches.
\end{enumerate} 
\end{enumerate} 
\begin{figure}
\centering
\begin{subfigure}{.33\textwidth}
  \centering
  \includegraphics[width=1\linewidth]{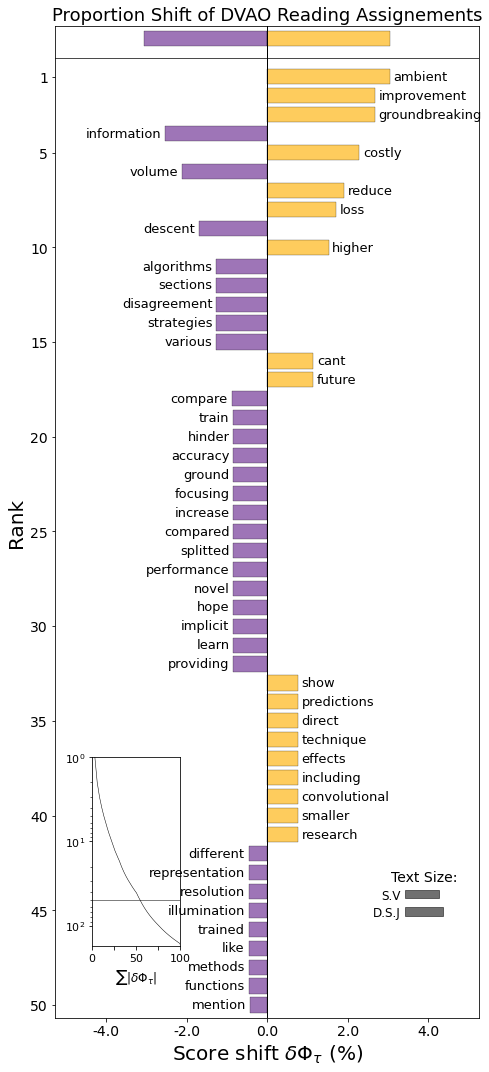}
  \caption{Frequency Comparision}
  \label{fig:a10freq}
\end{subfigure}%
\begin{subfigure}{.33\textwidth}
  \centering
  \includegraphics[width=1\linewidth]{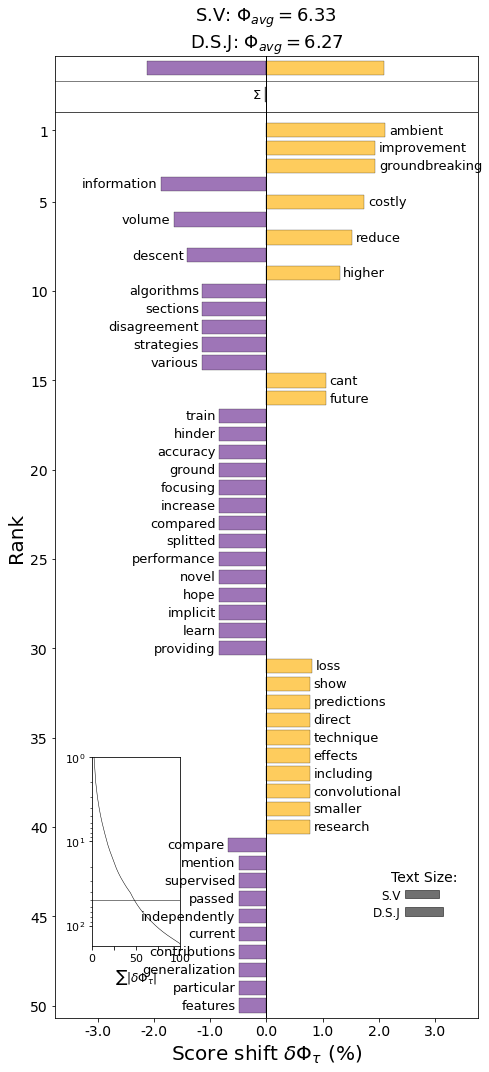}
  \caption{Entropy Comparision}
  \label{fig:a10entropy}
\end{subfigure}
\begin{subfigure}{.33\textwidth}
  \centering
  \includegraphics[width=1\linewidth]{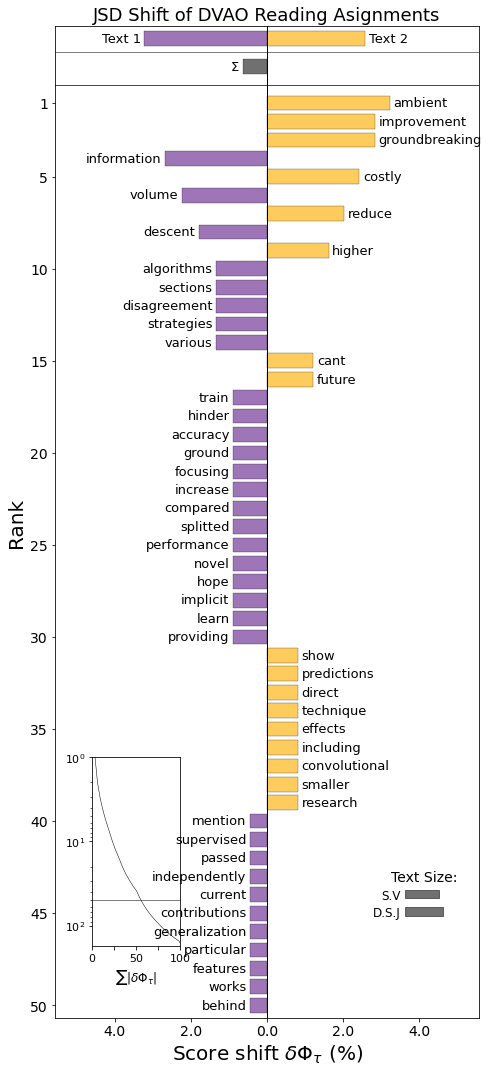}
  \caption{Divergence Comparison}
  \label{fig:a10kld}
\end{subfigure}
\caption{Word Shift Graphs for Reading Assignments}
\label{fig:A10graphs}
\end{figure}
\subsubsection{ECS289H: Reading Assignment 10's Summary by Divjeet Singh Jas and Suvi Varshney}
To test our model on academic reviews, we thought the best reviews to compare would be our own reviews. The word shift graphs are given in \ref{fig:A10graphs}

\begin{enumerate}
        \item Frequency comparison word shift graph:
        In Figure \ref{fig:a10freq}, the comparison of the Reading assignments is given, Suvi Varshney (in purple) and Divjeet Singh Jas (in yellow). The first thing we notice is the small difference in the corpus size. We also notice that the top 50 sentiment-carrying words, ranked by frequency, have an equal number of words from both the assignments, indicating that both the assignments are equally diverse. We can see that while Divjeet's assignment has more positive sentiment words like \textit{'improvement'}, \textit{'groundbreaking'}, etc. Suvi's assignment has more negative sentiment words like \textit{'disagreement'}, \textit{'descent'}, etc. This indicates that Divjeet's assignment paints the Reading in a more positive light, while Suvi's assignment is more critical of it. Since the corpus size is small, we can also notice that most of the difference is explained by the top 50 words combined, evident by the cumulative contribution plot on the bottom left.
\\
\item Entropy comparison word shift graph:
In Figure \ref{fig:a10entropy}, we can see which words are relatively surprising in which assignment. The words \textit{‘costly’}, \textit{‘ambient’} are much more surprising for Divjeet, while the words, \textit{‘Volume’}, \textit{‘Algorithms’} are much more surprising in Suvi's speech. The top bar shows the cumulative sum of $\delta\mathcal{H}$ and tells us that both the assignments are equally unpredictable.
\\
\item Divergence comparison word shift graph: In Figure \ref{fig:a10kld}, we can see that the word \textit{‘reduce’} in Divjeet’s assignment is surprising with respect to the reference assignment of Suvi. Similarly, the word \textit{‘information’} in Suvi’s assignment is surprising compared to the reference i.e Divjeet’s assignment.
\\
\end{enumerate} 
We also tested our pipeline on much more rigorously on other journal reviews and movie reviews, the results of which we have reserved for the presentation due to limited space.
\begin{figure}[h!]
\centering
\begin{subfigure}{.33\textwidth}
  \centering
  \includegraphics[width=1\linewidth]{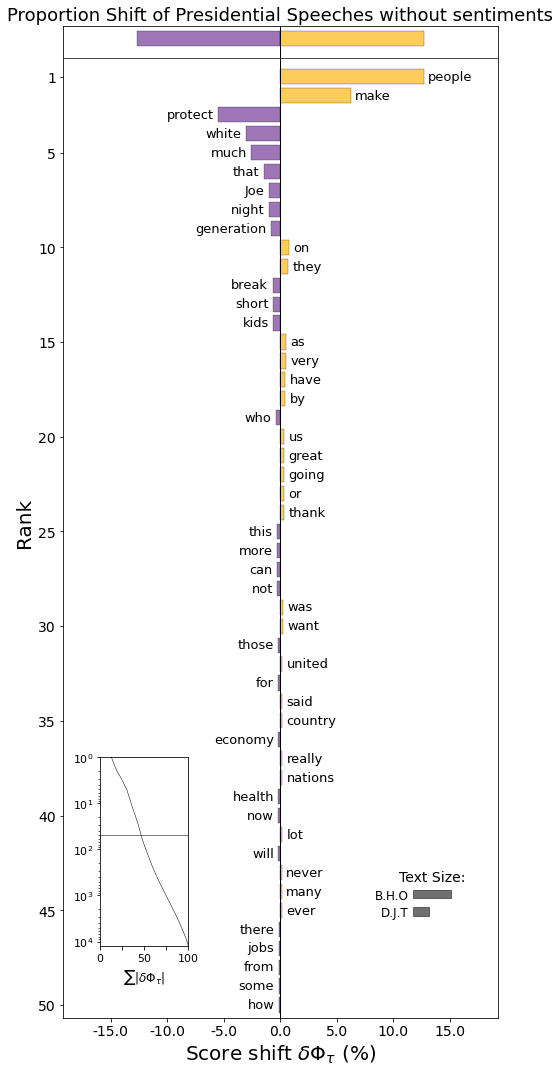}
  \caption{Frequency Comparision}
  \label{fig:nsfreq}
\end{subfigure}%
\begin{subfigure}{.33\textwidth}
  \centering
  \includegraphics[width=1\linewidth]{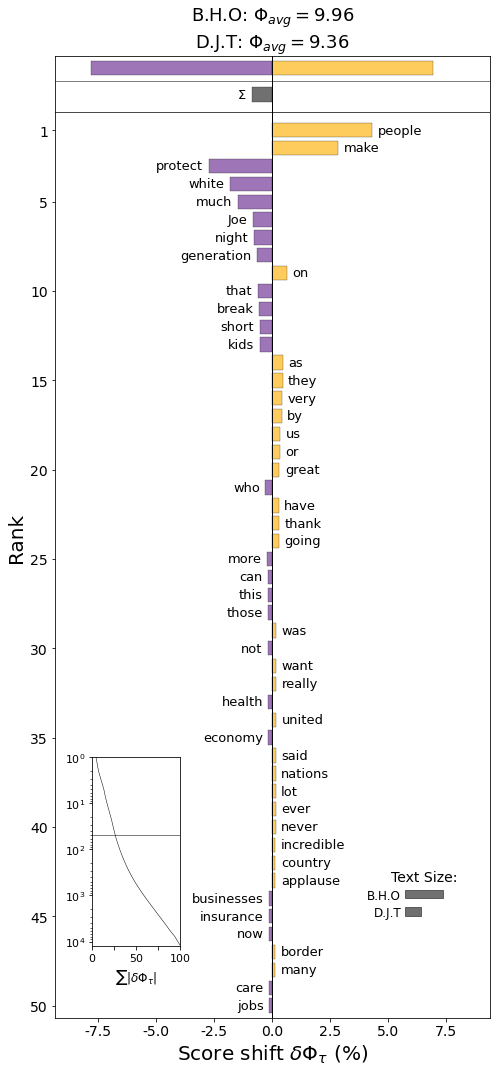}
  \caption{Entropy Comparision}
  \label{fig:nsentropy}
\end{subfigure}
\begin{subfigure}{.33\textwidth}
  \centering
  \includegraphics[width=1\linewidth]{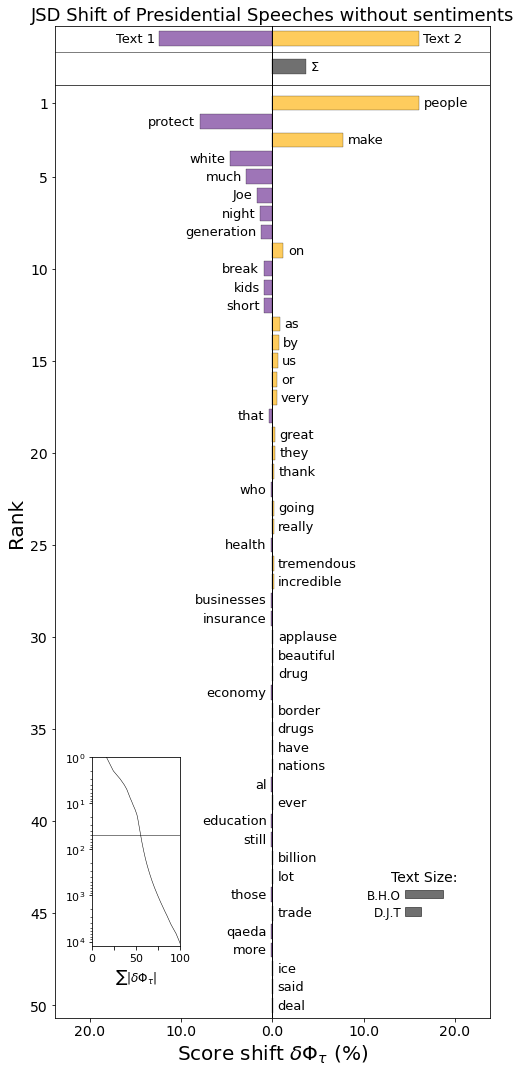}
  \caption{Divergence Comparison}
  \label{fig:nskld}
\end{subfigure}
\caption{Word Shift Graphs without Sentiment Analysis model}
\label{fig:NoSenti}
\end{figure}
\subsection{Comparison with existing work}
To highlight our contribution and the uniqueness of our pipeline, we present the result of word shift graphs on topics (not sentiment words) and try to see if they convey as much meaning. This comparison of speeches of Barack H. Obama (B.H.O in purple) and Donald J. Trump (D.J.T in yellow) is given.is given in Figure \ref{fig:NoSenti}. 
\\
As we can see in the frequency comparison word shift graph \ref{fig:nsfreq}, we have all kinds of words and not just sentiment words, they tell us with what frequency they appear, but do not convey anything about the sentiment or the intent of the corpus. Words such as \textit{'people', 'make', 'white', 'much', 'kids'} are not of much help to see the context. We can also see the cumulative contribution plot is not smooth, indicating that taking more words does not contribute equally in explaining the difference. Some words explain much more difference than others, which are probably sentiment-carrying words. The same trend is continued in the entropy and divergence word shift graphs in figure \ref{fig:nsentropy} and figure \ref{fig:nskld}

\section{Discussion, Limitation and Future Work}
This research is valuable for anyone who wants to analyze huge corpora in very little time. For example, a dataset containing critiques of an academic journal by different reviewers across the world. First, our model will generate a high-level view using the word cloud graph. This will provide the user with the topmost labeled sentiment-carrying words for every node. Along with the sentiment carrying common words among the different nodes. After this, the users can choose any 2 nodes they want to further analyze in-depth. This can be done by the word shift graphs generated by our model.

Since our current research is a proof of concept there is a need to further improve it and work on some of its limitations. Since we are calculating the sentiment of every word, our process takes a lot of time due to the huge corpus. This is a bottleneck in our current process. In the future, this can be solved by running the processes in parallel threads. Our current implementation is a proof of concept focusing on the presidential dataset. In the future, we can make this a more generalized approach for different datasets. Also currently, there is no support to add a dataset for evaluation in the world cloud graph. Our future implementation will evaluate the entire dataset and generate all the graphs.
\section{Documentation and Access}
\noindent The code base for this project is available at GitHub at \cite{git}.

\nocite{*}
\bibliography{main}

\begin{thebibliography}{14}
\newcommand{\enquote}[1]{``#1''}
\providecommand{\natexlab}[1]{#1}
\providecommand{\url}[1]{\texttt{#1}}
\providecommand{\urlprefix}{URL }
\expandafter\ifx\csname urlstyle\endcsname\relax
  \providecommand{\doi}[1]{\discretionary{}{}{}https://doi.org/#1}\else
  \providecommand{\doi}[1]{\discretionary{}{}{}\urlstyle{rm}\url{https://doi.org/#1}}\fi

\bibitem[{dat(2012)}]{dataset}
\enquote{\textit{Presidential Speech Dataset},}
  \url{https://millercenter.org/the-presidency/presidential-speeches }, 2012.

\bibitem[{Mikolov et~al.(2010)Mikolov, Karafi{\'a}t, Burget, Cernock{\'y}, and
  Khudanpur}]{Mikolov}
Mikolov, T., Karafi{\'a}t, M., Burget, L., Cernock{\'y}, J.~H., and Khudanpur,
  S., \enquote{Recurrent neural network based language model,}
  \emph{INTERSPEECH}, 2010.

\bibitem[{Hochreiter and Schmidhuber(1997)}]{LSTM}
Hochreiter, S., and Schmidhuber, J., \enquote{Long Short-Term Memory,} Vol.~9,
  No.~8, 1997, p. 1735–1780.
\newblock \doi{10.1162/neco.1997.9.8.1735},
  \urlprefix\url{https://doi.org/10.1162/neco.1997.9.8.1735}.

\bibitem[{Monroe et~al.(2008)Monroe, Colaresi, and Quinn}]{3}
Monroe, B.~L., Colaresi, M.~P., and Quinn, K.~M., \enquote{Fightin' Words:
  Lexical Feature Selection and Evaluation for Identifying the Content of
  Political Conflict,} \emph{Political Analysis}, Vol.~16, No.~4, 2008, pp.
  372--403.
\newblock
  \urlprefix\url{https://EconPapers.repec.org/RePEc:cup:polals:v:16:y:2008:i:04:p:372-403_00}.

\bibitem[{Bitvai and Cohn(2015)}]{bitvai}
Bitvai, Z., and Cohn, T., \enquote{Non-Linear Text Regression with a Deep
  Convolutional Neural Network,} \emph{Proceedings of the 53rd Annual Meeting
  of the Association for Computational Linguistics and the 7th International
  Joint Conference on Natural Language Processing (Volume 2: Short Papers)},
  Association for Computational Linguistics, Beijing, China, 2015, pp.
  180--185.
\newblock \doi{10.3115/v1/P15-2030},
  \urlprefix\url{https://aclanthology.org/P15-2030}.

\bibitem[{Coppersmith and Kelly(2014)}]{Coppersmith}
Coppersmith, G.~A., and Kelly, E., \enquote{Dynamic Wordclouds and Vennclouds
  for Exploratory Data Analysis,} 2014.

\bibitem[{Mike~Bostock and Ericson(2012)}]{bostock}
Mike~Bostock, S.~C., and Ericson, M., \enquote{At the national conventions, the
  words they used,}
  \url{https://archive.nytimes.com/www.nytimes.com/interactive/2012/09/06/us/politics/convention-word-counts.html?},
  2012.

\bibitem[{Shannon(1948)}]{shannon}
Shannon, C.~E., \enquote{A Mathematical Theory of Communication,} \emph{The
  Bell System Technical Journal}, Vol.~27, 1948, pp. 379--423.
\newblock
  \urlprefix\url{http://plan9.bell-labs.com/cm/ms/what/shannonday/shannon1948.pdf}.

\bibitem[{Sun et~al.(2019)Sun, Huang, and Qiu}]{sun-etal-2019-utilizing}
Sun, C., Huang, L., and Qiu, X., \enquote{Utilizing {BERT} for Aspect-Based
  Sentiment Analysis via Constructing Auxiliary Sentence,} \emph{Proceedings of
  the 2019 Conference of the North {A}merican Chapter of the Association for
  Computational Linguistics: Human Language Technologies, Volume 1 (Long and
  Short Papers)}, Association for Computational Linguistics, Minneapolis,
  Minnesota, 2019, pp. 380--385.
\newblock \urlprefix\url{https://www.aclweb.org/anthology/N19-1035}.

\bibitem[{dat(2017)}]{dataset-senti}
\enquote{\textit{SemEval-2014 ABSA Restaurant Reviews},}
  \url{http://metashare.ilsp.gr:8080/repository/browse/semeval-2014-absa-restaurant-reviews-train-data/479d18c0625011e38685842b2b6a04d72cb57ba6c07743b9879d1a04e72185b8/
  }, 2017.

\bibitem[{Gallagher et~al.(2021)Gallagher, Frank, Mitchell, Schwartz, Reagan,
  Danforth, and Dodds}]{Shifterator}
Gallagher, R.~J., Frank, M.~R., Mitchell, L., Schwartz, A.~J., Reagan, A.~J.,
  Danforth, C.~M., and Dodds, P.~S., \enquote{Generalized word shift graphs: a
  method for visualizing and explaining pairwise comparisons between texts,}
  \emph{EPJ Data Science}, Vol.~10, No.~1, 2021.
\newblock \doi{10.1140/epjds/s13688-021-00260-3},
  \urlprefix\url{http://dx.doi.org/10.1140/epjds/s13688-021-00260-3}.

\bibitem[{git(2022)}]{git}
\enquote{\textit{GitHub Repository},}
  \url{https://github.com/suvivarshney002/CompText}, 2022.

\bibitem[{int(2021)}]{interfaces}
\enquote{\textit{IMDB Dataset},} \url{https://www.imdb.com/interfaces/}, 2021.

\bibitem[{Knittel et~al.(2021)Knittel, Koch, Tang, Chen, Wu, Liu, and
  Ertl}]{DBLP:journals/corr/abs-2108-03052}
Knittel, J., Koch, S., Tang, T., Chen, W., Wu, Y., Liu, S., and Ertl, T.,
  \enquote{Real-Time Visual Analysis of High-Volume Social Media Posts,}
  \emph{CoRR}, Vol. abs/2108.03052, 2021.
\newblock \urlprefix\url{https://arxiv.org/abs/2108.03052}.

\end{thebibliography}

\end{document}